\documentclass[letterpaper, 10 pt,conference,hidelinks]{IEEEtran}

\usepackage[top=1in, bottom=0.75in, left=0.75in, right=0.75in]{geometry}

\IEEEoverridecommandlockouts
\usepackage{cite}
\usepackage{amsmath,amssymb,amsfonts}
\usepackage{algorithmic}
\usepackage{graphicx}
\usepackage{textcomp}
\usepackage{xcolor}
\usepackage{amsmath}
\usepackage{hyperref}
\usepackage{subfigure}
\usepackage{multirow}
\usepackage{booktabs}

\usepackage[normalem]{ulem}

\usepackage{microtype}  

\makeatletter
\newcommand{\linebreakand}{%
  \end{@IEEEauthorhalign}
  \hfill\mbox{}\par
  \mbox{}\hfill\begin{@IEEEauthorhalign}
}
\makeatother

\def\BibTeX{{\rm B\kern-.05em{\sc i\kern-.025em b}\kern-.08em
    T\kern-.1667em\lower.7ex\hbox{E}\kern-.125emX}}

\usepackage{siunitx}

\DeclareRobustCommand*{\IEEEauthorrefmark}[1]{%
    \raisebox{0pt}[0pt][0pt]{\textsuperscript{\footnotesize\ensuremath{#1}}}}

\begin{document}
\sloppy
\title{
\textbf{Simulated Cortical Magnification Supports Self-Supervised Object Learning}

\thanks{\textcolor{black}{ZY thanks the Xidian-FIAS International Joint Research Center for funding. This research was supported by “The Adaptive Mind” funded by the Excellence Program of the Hessian Ministry of Higher Education, Research, Science and the Arts, Germany, and by the Deutsche Forschungsgemeinschaft (DFG project “Abstract REpresentations in Neural Architectures (ARENA)”). This work was funded by NIH R01HD074601 and R01HD093792 to CY. JT was supported by the Johanna Quandt foundation. We thank the members of the Developing Intelligence Lab at UT Austin and participating families for their contributions. We gratefully acknowledge support from Goethe University (NHR Center NHR@SW) for providing some of the computing and data-processing resources needed for this work.}
}
}


\author{
    \IEEEauthorblockN{Zhengyang Yu\IEEEauthorrefmark{1,2}, 
                      Arthur Aubret\IEEEauthorrefmark{1,2}, 
                      Chen Yu\IEEEauthorrefmark{3}
                      and
                      Jochen Triesch\IEEEauthorrefmark{1}}
                      
    \IEEEauthorblockA{\IEEEauthorrefmark{1} \textit{Frankfurt Institute for Advanced Studies}, Frankfurt am Main, Germany}
    \IEEEauthorblockA{\IEEEauthorrefmark{2} \textit{Xidian-FIAS International Joint Research Center}, Frankfurt am Main, Germany}
    \IEEEauthorblockA{\IEEEauthorrefmark{3} \textit{Department of Psychology, University of Texas at Austin}, Austin, USA}
    \IEEEauthorblockA{\{zhyu, aubret, triesch\}@fias.uni-frankfurt.de, chen.yu@austin.utexas.edu}
}

\maketitle


\begin{abstract}
Recent self-supervised learning models simulate the development of semantic object representations by training on visual experience similar to that of toddlers. However, these models ignore the foveated nature of human vision with high/low resolution in the center/periphery of the visual field. Here, we investigate the role of this varying resolution in the development of object representations. We leverage two datasets of egocentric videos that capture the visual experience of humans during interactions with objects. We apply models of human foveation and cortical magnification to modify these inputs, such that the visual content becomes less distinct towards the periphery. The resulting sequences are used to train two bio-inspired self-supervised learning models that implement a time-based learning objective. Our results show that modeling aspects of foveated vision improves the quality of the learned object representations in this setting. Our analysis suggests that this improvement comes from making objects appear bigger and inducing a better trade-off between central and peripheral visual information. Overall, this work takes a step towards making models of humans' learning of visual representations more realistic and performant.
\end{abstract}


\section{Introduction}

Recent self-supervised learning through time (SSLTT) models aim to emulate the biological development of semantic object representations by training on visual experiences similar to humans' with a bio-inspired learning mechanism \cite{schneider2021contrastive,aubret2022time}. However, these models often overlook the high/low resolution of human vision in the central/peripheral field of view. The retina has the highest density of receptors in the fovea, which therefore samples the central part of the visual field at a high resolution. In contrast, receptor density in the periphery is low, leading to a low sampling resolution \cite{banks1991peripheral}. This differential sampling of information at the level of the retina, referred to as Foveation, is also reflected in the primary visual cortex, where the representation of the fovea takes up relatively much more cortical space than the representation of the periphery. This is usually measured as the so-called Cortical Magnification, which describes how many square millimeters of cortical surface area are devoted to a certain portion of the visual field sampled at different retinal locations \cite{costantino2022feedback}. In human vision, Foveation and Cortical Magnification are two sides of the same coin, reflecting the space-variant sampling of visual information.

This spatial disparity in visual processing may play an important role in shaping visual representations. Specifically, it may govern the trade-off between \textcolor{black}{the extraction} of features from central and peripheral vision. When humans direct their gaze towards an object, this increases the amount of processing dedicated to the object versus the background. This is especially important, as previous studies have found that models trained on the visual experience of toddlers tend to focus more on the background rather than on the object present in the foreground \cite{orhan2024learning,aubret2022toddler}.

In this study, we investigate whether and how Foveation and Cortical Magnification impact the learning of visual representations in a model of children's learning of object representations. We leverage a real-world dataset capturing the visual experience of toddlers as they interact with toys, providing authentic gaze fixation data \cite{bambach2018toddler}. Additionally, we use the CORe50 dataset, which presents typical object observations from an adult perspective \cite{lomonaco2017core50}. We modify these videos by applying two image augmentations that simulate Foveation and Cortical Magnification \textcolor{black}{and their combination.} \textcolor{black}{We} simulate human learning by training bio-inspired learning models on the resulting visual sequences. 

Our results indicate that the Cortical Magnification supports the learning of object representations. Our analysis suggests that Cortical Magnification induces a better trade-off between central and peripheral visual information, which benefits visual learning. These findings demonstrate the value of \textcolor{black}{the Cortical Magnification augmentation for} developing strong object representations.



\section{Related works}
\subsection{Self-Supervised learning through time}
The biological principle of temporal slowness states that biological systems extract similar semantic representations for close-in-time visual inputs \cite{wiskott2002slow,li2008unsupervised}. Early studies demonstrated that learning based on temporal slowness could extract representations of simple patterns that are unaffected by changes in position, size, and rotation \cite{wiskott2002slow}. Other research applied this principle to learn view-invariant object recognition \cite{stringer2006learning,franzius2011invariant,schneider2021contrastive}. Recent advances in self-supervised learning have enabled the scaling of this temporal slowness principle to large sets of uncurated images of objects \cite{parthasarathy2022self}, a method known as SSLTT \cite{aubret2022time}. On the machine learning side, SSLTT has been shown to enhance category recognition \cite{aubret2024self,aubret2022time}, view-invariant object instance recognition \cite{schneider2021contrastive}, and alignment with human representations \cite{parthasarathy2023self}. From the perspective of cognitive modeling, SSLTT helps shape human-like inter-object semantic similarities \cite{aubret2024learning} and integrates effectively with visuo-language SSL to model object learning during dyadic play \cite{schaumloffel2023caregiver}. Despite the impressive advances of SSLTT, these models do not simulate Cortical Magnification, a process that may be important for visual object learning. 



\subsection{Simulations of Foveation and Cortical Magnification}



\cite{lukanov2021biologically} proposed an end-to-end neural model to simulate foveal-peripheral vision, inspired by retino-cortical mapping in primates and humans. This model introduces an efficient sampling technique that compresses the visual signal, allowing a small portion of the scene to be perceived in high resolution while maintaining a large field of view in low resolution. In computational models, this allocation of visual attention and resolution plays a key role in optimizing both the computational efficiency and the quality of visual perception. While both Foveation and Cortical Magnification have been studied in supervised and self-supervised learning \cite{bambach2018toddler,lukanov2021biologically,wang2021use}, there is a lack of research on their role within SSLTT. Recent works have proposed learning representations on the visual part that corresponds to central vision \cite{yu2024active,schaumloffel2025human}. However, they neglect the low resolution of peripheral vision. Another work studied the impact of Foveation when training with few objects and exhibited that SSLTT tends to extract background versus object information \cite{aubret2022toddler}; however, they did not use humans' real-world visual experience and did not model Cortical Magnification. In \sectionautorefname~\ref{sec:expe}, we find that Cortical Magnification has a different effect compared to Foveation.



\section{Methods}
\begin{figure}[t]
    \centering
    \includegraphics[width=\linewidth]{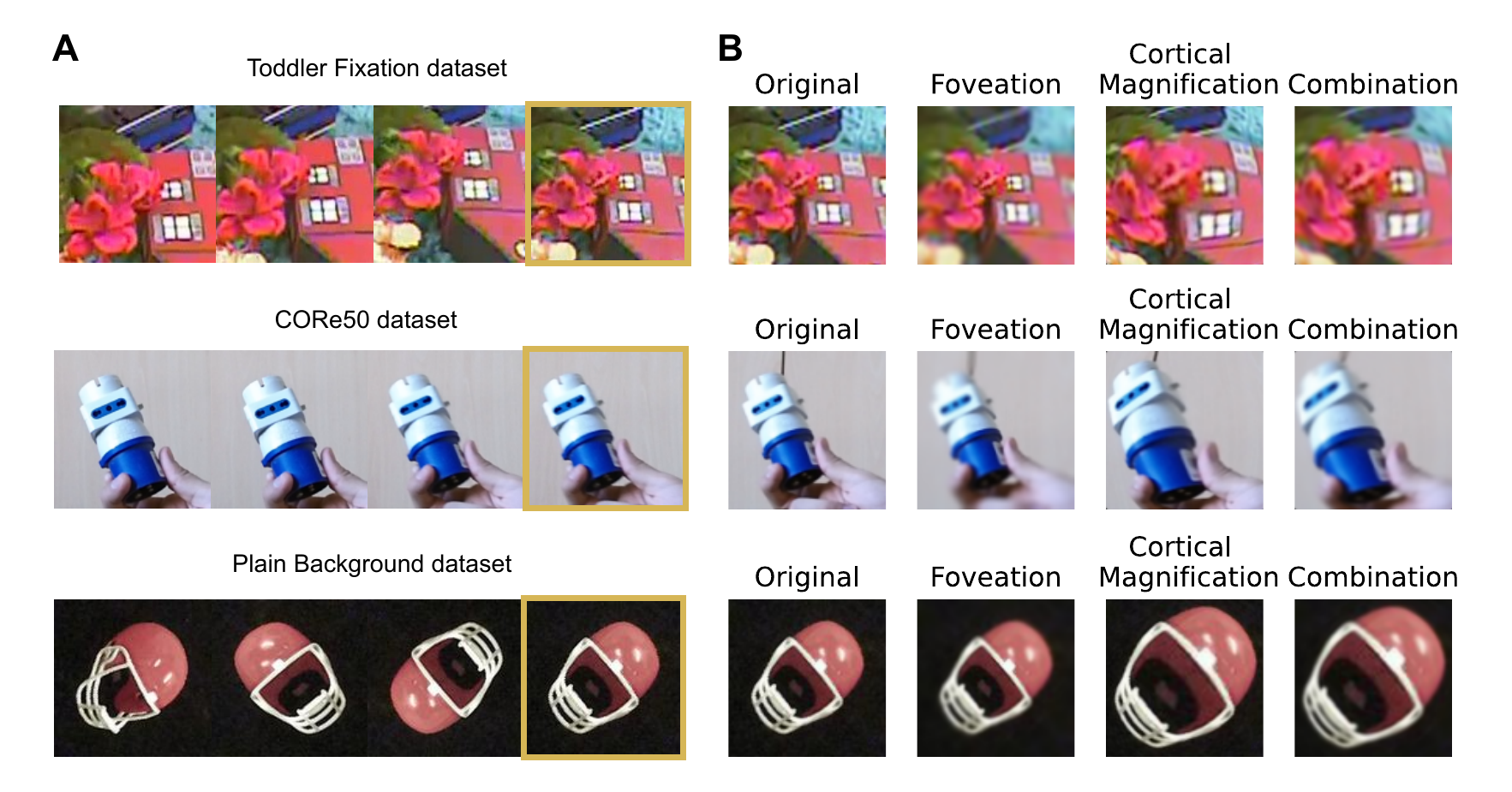}
    \caption{Comparison of biological augmentations across three datasets: Toddler Fixation \cite{bambach2018toddler}, CORe50 \cite{lomonaco2017core50}, and Plain Background \cite{bambach2018toddler}. {\bf (A)} presents examples from the three datasets while {\bf (B)} shows their augmented versions. We showcase different processing augmentations of the image enclosed in the yellow box: None, Foveation, Cortical Magnification, and the Combination. 
    }
    \label{fig:dataset}
\end{figure}

\begin{figure}[t]
    \centering
    \includegraphics[width=\linewidth]{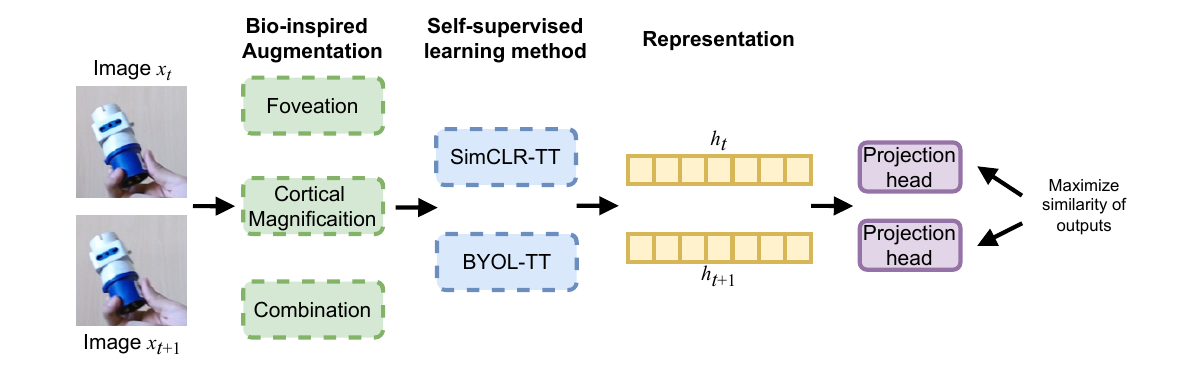}
    \caption{Overview of our framework. Two consecutive video frames, $x_t$ and $x_{t+1}$, are first processed by one of the bio-inspired augmentation methods \textcolor{black}{or their combination.} The augmented images are then fed into a self-supervised learning method, chosen from SimCLR-TT or BYOL-TT, to extract representations $h_t$ and $h_{t+1}$. The training objective maximizes the similarity between the representations of temporally adjacent frames. In this framework, dashed boxes indicate that only one augmentation method and one self-supervised learning method are selected during each training run.}
    \label{fig:overview}
\end{figure}

We aim to investigate the impact of the varying resolution of human vision on the learning of object representations. We leverage two real-world datasets showing natural interactions with objects from a first-person perspective. Then, we modify the visual sequences with a model of Foveation and one of Cortical Magnification. Finally, we train a bio-inspired self-supervised learning model based on temporal slowness on the resulting visual sequences. \figureautorefname~\ref{fig:dataset}A provides examples of the dataset utilized in this study. An overview of our framework is shown in \figureautorefname~\ref{fig:overview}.


\subsection{Datasets}\label{sec:datasets}

\textbf{Toddler Fixation dataset.} This dataset shows videos that capture the visual experience of toddlers during play sessions, derived from head-cameras and eye trackers \cite{bambach2018toddler,yu2024active}. To simulate the visual experience of toddlers, we follow the procedure previously introduced in \cite{yu2024active} and extract from the original videos \cite{bambach2018toddler} a $128\times128$ crop centered on the fixation point of toddlers. This corresponds to a visual angle of $\SI{14}{\degree} \times \SI{14}{\degree}$. The dataset comprises 559,522 images and shows interactions with a total of 24 toys.

\textbf{CORe50 dataset.} The CORe50 dataset is specifically designed for continual learning and object recognition tasks from a first-person perspective \cite{lomonaco2017core50}. It contains 50 object categories, each photographed in 11 different sessions, covering variations in perspective, lighting conditions, and backgrounds. In this study, the full dataset consists of 164,866 $128\times128$ RGB-D images: $\text{11 sessions} \times \text{50 objects} \times \text{(around 300) frames per session}$.

\textbf{Plain background dataset.} 
This dataset shows the same objects as the Toddler Fixation dataset, but in front of a black background. It contains 128 viewpoints, capturing each object from various angles and distances for 1,536 images. Each image in this dataset displays a complete object against a black background, ensuring visual isolation from external distractions.

Here, the purpose of the Toddler Fixation dataset is to investigate the role of Foveation and Cortical Magnification in Self-supervised object learning under real-world viewing conditions and fixation points. In contrast, the CORe50 dataset provides a scenario where both the entire object and the simulated fixation point are centrally positioned. Finally, the Plain Background dataset further simplifies the conditions to an ideal case, allowing for an exploration of how these two mechanisms work when objects are distinguishable from the background and centrally located.

\subsection{Foveation and Cortical Magnification}

To simulate the varying resolution of human vision, we apply two biologically inspired augmentations to train and test images, namely Foveation and Cortical Magnification.  

\textbf{Foveation augmentation.} A simple approach to modeling high-resolution central vision and low resolution in the periphery is to apply a spatially varying blur \cite{geisler1998real}. Thus, we follow a previous method \cite{wang2021use} to implement a Foveation transform $\mathcal{F}$. Given a fixation point, the pixels in the image are divided into belts of similar eccentricity, represented by masks $M_i$. Then, the image $I$ is convolved with Gaussian kernels $K(\sigma_i)$ of different standard deviation $\sigma_i$, corresponding to the degree of blur at different eccentricities. Finally, these blurred images $I * K(\sigma_i)$ are blended into one with the belt-shaped masks. Concretely, we define 
\begin{equation}
\mathcal{F}(I)=\sum_i^{N_b} M_i \cdot\left[I * K\left(\sigma_i\right)\right], \quad \sum_i^{N_b} M_i[i, j]=1,
\end{equation}
and display the visual outcome in \figureautorefname~\ref{fig:dataset}.
\textbf{Cortical Magnification augmentation.} Due to the different densities of retina ganglion cells in central and peripheral vision \cite{wassle1989cortical}, the cortical area corresponding to the unit area of the retinal image varies as a function of eccentricity \cite{van1984visual}. Thus, we can think of ``cortical images'' as a warped version of the retinal image \cite{yates2021beyond}. 

The Cortical Magnification Factor (CMF) represents the cortical surface distance between two points representing visual field positions and is an inverse function of eccentricity \cite{harvey2011relationship}.

We follow the implementation of a previous study \cite{wang2021use}. First, we assume the CMF is constant close to the fovea since the data there is usually lacking. Therefore, the CMF will be represented as a piecewise function.
\begin{align}
C M F(r)= \begin{cases}C & r<r_{\text {fov }} \\ \frac{C\left(r+r_{\text {fov }}\right)}{r+K} & r \geq r_{\text {fov }}\end{cases} \; {\rm and}
\end{align}


Then, we transform the retinal eccentricity $e(r)$ to cortical radial distance by integrating the reciprocal of
the CMF. Thus, $e(r)$ is a piecewise linear (foveal) or quadratic (peripheral) function of the cortical radius $r$: 
\begin{equation}
\begin{aligned}
e(r)&=\int_0^r \frac{1}{C M F(\rho)} d \rho\\&=\frac{1}{C} \begin{cases}r ; & r<r_{\text {fov }} \\ \frac{(r+K)^2}{2\left(r_{\text {fov }}+K\right)}+\frac{r_{\text {fov }}-K}{2} ; & r \geq r_{\text {fov }}\end{cases} \; .
\end{aligned}
\end{equation}

We assume an isotropic transform from retinal images to cortical images, so we can use this radial transform to build the 2d change of coordinates $(r, \theta) \xrightarrow{} (e(r), \theta)$. Note that the constant $C=1$ controls the scaling of the map, which affects the area of the image covered by the sampling grid. In addition, the parameter $r_{fov}$ controls the radial distortion area of linear sampling, while $K$ controls the degree of distortion in the periphery. 
Unless stated otherwise,  we set $ r_{\text {fov }} = 20$ and $K = 20$ (both measured in pixels). We also propose to apply simultaneously Foveation and Cortical Magnification; the effect can be visualized in \figureautorefname~\ref{fig:dataset}B.




\subsection{Learning models}
To model visual learning in toddlers, we leverage self-supervised models that implement the slowness principle. This well-established principle of biological learning \cite{li2010unsupervised} states that biological systems learn similar visual representations for visual inputs that are close in time. We use two SSLTT models introduced in \cite{schneider2021contrastive}, called SimCLR-TT and BYOL-TT. 

\textbf{SimCLR-TT.} This model is based on the state-of-the-art SimCLR method \cite{chen2020simple}. SimCLR-TT samples an image $x_i$ at time $i$ and a temporally close image $x_j$ as one positive pair and computes their respective embeddings $z_i$, $z_j$ with a deep neural network \cite{schneider2021contrastive} (e.g.\ a ResNet). Typically, the time interval between two adjacent images is the reciprocal of the camera's FPS. Then, SimCLR-TT minimizes
\begin{equation}
\mathcal{L}\left(z_i,z_j\right)=-\log \frac{\exp \left(\operatorname{sim}\left(z_i, z_j\right) / \tau\right)}{\sum_{z_k \in \mathcal{B}, k \neq i}^{}\left[\exp \left(\operatorname{sim}\left(z_i, z_{k}\right) / \tau\right)\right]},
\end{equation}
where $\mathcal{B}$ is a minibatch, ${\rm sim}(\cdot)$ is the cosine similarity and $\tau$ is the temperature hyper-parameter. Thus, SimCLR-TT maximizes the similarity between temporally close representations (numerator) while keeping all representations dissimilar from each other (denominator). 


\textbf{BYOL-TT.} We also consider BYOL-TT, another self-supervised learning method that applies the principle of temporal slowness \cite{grill2020bootstrap,schneider2021contrastive}.

\subsection{Training and evaluation}
We train and evaluate two SSLTT models on three different datasets to investigate whether Foveation and Cortical Magnification contribute to improved learning object representations in self-supervised learning models.

We chose ResNet18 (\(\sim11.7\)M parameters, \(\sim0.59\) GFLOPs) and ResNet50 (\(\sim25.6\)M parameters, \(\sim1.34\) GFLOPs) for their simplicity and proven effectiveness in contrastive and time-based self-supervised learning \cite{schneider2021contrastive}. Moreover, their depth and residual connections have been suggested to partially approximate recurrent processing \cite{liao2016bridging}, providing a practical balance between biological plausibility and simplicity in our framework. \textcolor{black}{All ResNet backbones in our experiments are trained entirely from scratch, without using any pretrained weights.}

We run all experiments with three random seeds and train the models for 100 epochs using ResNet18 and ResNet50 as visual encoders. 
The initial learning rate of $10^{-2}$ and weight decay of $10^{-4}$ follow \cite{yu2024active} and were verified through preliminary tuning. We set the SimCLR temperature to $0.08$ and use a batch size of 256, also consistent with recommended settings in \cite{yu2024active}.

We assess the quality of the learned representations by training a linear classifier on top of the learned representation (right after the average pooling layer) in a supervised fashion. This is the standard evaluation protocol in self-supervised visual learning \cite{chen2020simple}. Since the Toddler Fixation dataset does not contain labeled images, we train the linear classifier on the train split of the Objects Fixation dataset \cite{yu2024active} (manually extracted and labeled from the same videos). Then, we evaluate the object recognition accuracy of the classifier on the test split of the Objects Fixation dataset. For the CORe50 dataset, three of the eleven sessions (Session 3, 7, and 10) have been selected for testing, and the remaining 8 sessions are used for training, which aligns with the original setting in \cite{lomonaco2017core50}. For the Plain Background dataset, the entire dataset contains only 1536 images of 24 objects. We keep the last 20 $\%$ of the rotation angles as a test set, and use the rest of the dataset for training.

\section{Experiments}\label{sec:expe}

Our experiments aim to assess whether and how Foveation and Cortical magnification support object learning. We first compare how these image augmentations impact the learning of object representations. Then, we analyze how they impact the learning process and the effect of hyperparameters.

\subsection{Cortical Magnification supports learning objects}
In \tableautorefname~\ref{tab:acc_toddler} and \tableautorefname~\ref{tab:acc_core}, we show the results for the Toddler Fixation dataset and the CORe50 dataset, respectively. For comparison, we also include the performance of one supervised model to serve as an upper-bound reference. We observe that Cortical Magnification consistently improves or matches the performance of other methods. Especially for the CORe50 dataset, Cortical Magnification outperforms the absence of augmentations by up to $3\%$ of accuracy. In contrast, adding Foveation (Foveation \& Combination) leads only to a decrease (Toddler Fixation) or a slight increase (CORe50) in object recognition accuracy. We speculate that the ambivalent results of Foveation stem from the specific characteristics of the datasets. We observed that objects tend to appear larger in the Toddler Fixation dataset compared to the CORe50 dataset. As a result, applying Foveation may introduce more blur on parts of the objects on the Toddler Fixation dataset, which may hurt the recognition process. \textcolor{black}{Statistical tests indicate that, compared to the ``None'' condition, these bio-inspired augmentations yield significant improvements (Mann–Whitney U tests, $p < 0.05$, specifically $p \in [0.032, 0.045]$) for ResNet18 across all methods (Supervised, SimCLR-TT, BYOL-TT). Although numerical gains are also seen for ResNet50, they do not consistently reach statistical significance. Similar improvements are found on the Plain Background dataset. Overall, we conclude that modeling Cortical Magnification boosts the recognition of objects.}

\begin{table}[ht]
\caption{Test object accuracy of the models trained on the Toddler Fixation dataset with different settings of augmentations.}
\vspace{0.5em}
\tabcolsep=0.2cm
\centering
\resizebox{0.49\textwidth}{!}{
\renewcommand\arraystretch{1.8}
\begin{tabular}{cccccc}
\toprule
\multirow{2}{*}{Augmentation} & \multicolumn{1}{c}{Supervised} & \multicolumn{2}{c}{SimCLR-TT} & \multicolumn{2}{c}{BYOL-TT} \\ \cline{2-6} 
& ResNet18 & ResNet18    & ResNet50    & ResNet18   & ResNet50  \\ \hline
None
& $0.918\pm{0.009}$
& $0.863\pm{0.011}$
& $0.872\pm{0.005}$
& $0.823\pm{0.004}$ 
& $0.868\pm{0.014}$ \\\hline
Foveation
& $0.907\pm{0.012}$
& $0.852\pm{0.023}$ 
& $0.869\pm{0.016}$  
& $0.824\pm{0.024}$ 
& $0.857\pm{0.012}$ \\\hline
Cortical Magnification
& $\textbf{0.924}\pm{\textbf{0.014}}$
& $\textbf{0.869}\pm{\textbf{0.025}}$
& $0.871\pm{0.015}$ 
& $\textbf{0.835}\pm{\textbf{0.021}}$ 
& $\textbf{0.874}\pm{\textbf{0.018}}$  \\\hline
Combination
& $0.920\pm{0.013}$
& $0.865\pm{0.033}$ 
& $\textbf{0.876}\pm{\textbf{0.028}}$
& $0.831\pm{0.025}$ 
& $0.872\pm{0.031}$ \\
\bottomrule
\end{tabular}}
\label{tab:acc_toddler}
\end{table}

\begin{table}[ht]
\caption{Test object accuracy of the models trained on the CORe50 dataset with different settings of augmentations.}
\vspace{0.5em}
\tabcolsep=0.2cm
\centering
\resizebox{0.49\textwidth}{!}{
\renewcommand\arraystretch{1.8}
\begin{tabular}{cccccc}
\toprule
\multirow{2}{*}{Augmentation} & \multicolumn{1}{c}{Supervised} & \multicolumn{2}{c}{SimCLR-TT} & \multicolumn{2}{c}{BYOL-TT} \\ \cline{2-6} 
& ResNet18 & ResNet18    & ResNet50    & ResNet18   & ResNet50  \\ \hline
None
& $0.705\pm{0.013}$
& $0.608\pm{0.021}$
& $0.625\pm{0.030}$
& $0.597\pm{0.034}$ 
& $0.618\pm{0.026}$ \\\hline
Foveation
& $0.718\pm{0.021}$
& $0.613\pm{0.026}$ 
& $0.634\pm{0.050}$  
& $0.602\pm{0.033}$ 
& $0.625\pm{0.017}$ \\\hline
Cortical Magnification
& $0.729\pm{0.018}$
& $0.634\pm{0.028}$
& $0.656\pm{0.022}$ 
& $0.629\pm{0.034}$ 
& $\textbf{0.647}\pm{\textbf{0.016}}$  \\\hline
Combination
& $\textbf{0.737}\pm{\textbf{0.016}}$ 
& $\textbf{0.641}\pm{\textbf{0.019}}$ 
& $\textbf{0.658}\pm{\textbf{0.023}}$
& $\textbf{0.635}\pm{\textbf{0.013}}$ 
& $0.644\pm{0.024}$ \\
\bottomrule
\end{tabular}}
\label{tab:acc_core}
\end{table}




\subsection{Cortical Magnification makes objects appear bigger, which benefits learning}
\begin{table}[ht]
\caption{Test object accuracy of the models trained on the Plain Background dataset with different settings of augmentations. }
\vspace{0.5em}
\tabcolsep=0.2cm
\centering
\resizebox{0.49\textwidth}{!}{
\renewcommand\arraystretch{1.8}
\begin{tabular}{cccccc}
\toprule
\multirow{2}{*}{Augmentation} & \multicolumn{1}{c}{Supervised} & \multicolumn{2}{c}{SimCLR-TT} & \multicolumn{2}{c}{BYOL-TT} \\ \cline{2-6} 
& ResNet18 & ResNet18    & ResNet50    & ResNet18   & ResNet50  \\ \hline
None
& $0.968\pm{0.007}$
& $0.946\pm{0.018}$
& $0.951\pm{0.012}$
& $0.931\pm{0.016}$ 
& $0.940\pm{0.017}$ \\\hline
Foveation
& $0.961\pm{0.009}$
& $0.943\pm{0.013}$ 
& $0.947\pm{0.006}$  
& $0.928\pm{0.008}$ 
& $0.935\pm{0.019}$ \\\hline
Cortical Magnification
& $\textbf{0.978}\pm{\textbf{0.014}}$ 
& $\textbf{0.955}\pm{\textbf{0.026}}$ 
& $\textbf{0.961}\pm{\textbf{0.017}}$
& $\textbf{0.942}\pm{\textbf{0.014}}$ 
& $\textbf{0.948}\pm{\textbf{0.018}}$ \\\hline
Combination
& $0.973\pm{0.016}$
& $0.951\pm{0.011}$
& $0.957\pm{0.016}$ 
& $0.936\pm{0.012}$ 
& $0.944\pm{0.015}$  \\
\bottomrule
\end{tabular}}
\label{tab:acc_plain}
\end{table}

We propose two hypotheses to explain why Cortical Magnification performs better than the other augmentation settings. First, it may be that the augmentation attenuates the extraction of background features, which allows the model to focus more on object features \cite{schaumloffel2025human}. Second, it may make objects appear larger, a characteristic which is known to help object learning \cite{bambach2018toddler,aubret2022toddler}. To investigate these hypotheses, we train our models on the Plain Background dataset, which lacks background distractions. In \tableautorefname~\ref{tab:acc_plain}, we observe that Cortical Magnification \textcolor{black}{still outperforms} the other augmentation settings not only under self-supervised learning but also under supervised training. We conclude that parts of the performance boost stem from making objects appear larger.

\begin{figure}[ht]
  \centering
  \includegraphics[width=1\linewidth]{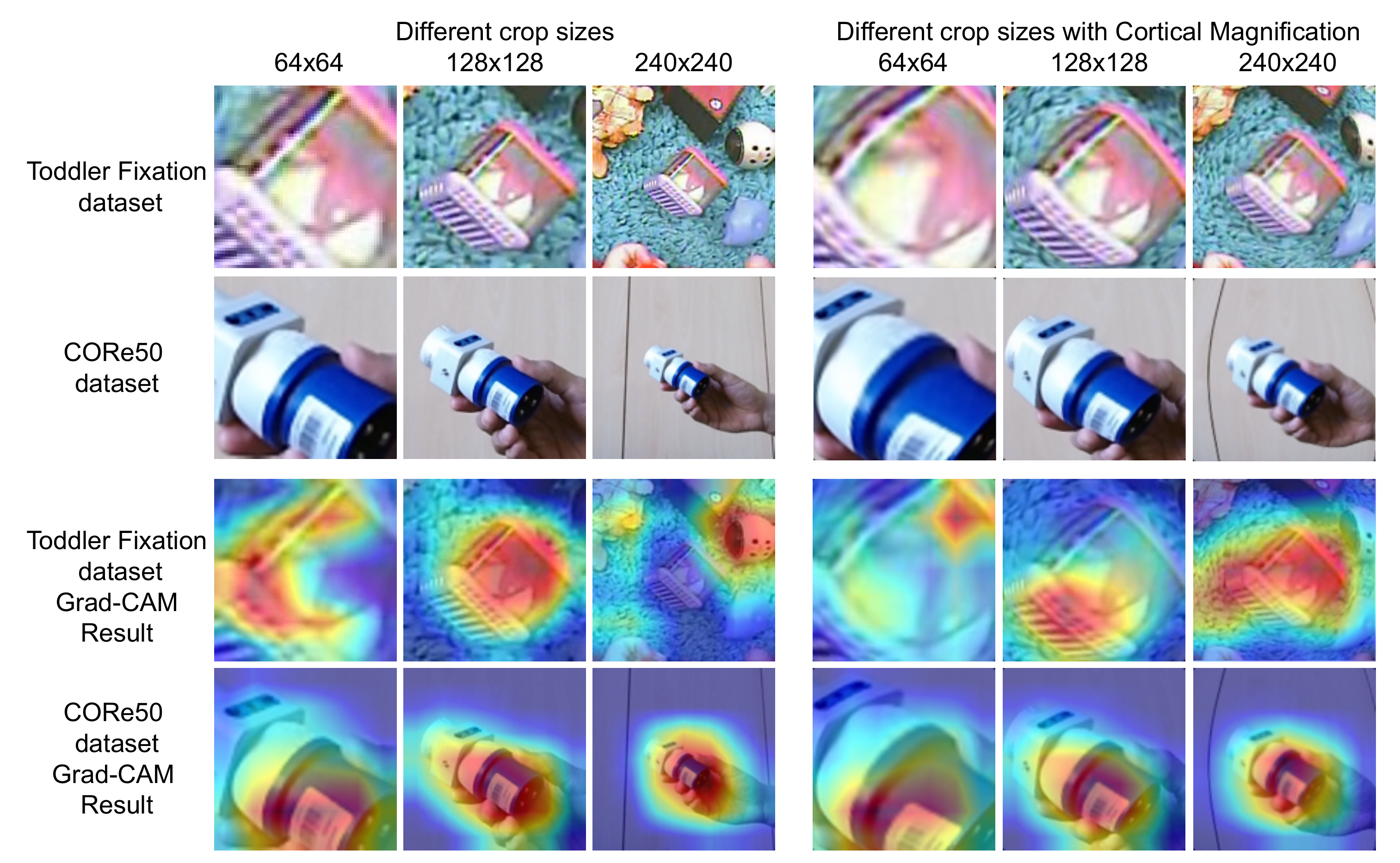}
  \caption{The visual effect of different crop sizes with or without Cortical Magnification. The last two rows show the attention map resulting from the application of Grad-CAM \cite{selvaraju2017grad}. We use SimCLR-TT with ResNet18 encoder and select one object from the Toddler Fixation dataset and one from the CORe50 dataset for visualization.}
  \label{fig:crop_example}
\end{figure}

To further analyze how making objects larger helps learning about objects, we compute in \figureautorefname~\ref{fig:crop_example} the attention map resulting from the application of Grad-CAM \cite{selvaraju2017grad}. With the Toddler Fixation dataset, we observe that at a small crop size of $64\times64$, the model's attention is dispersed due to partial object distortion. As the crop size increases to $240\times240$, Cortical Magnification makes the object appear larger relative to its original size, which helps the model focus more precisely on the relevant object. This suggests Cortical Magnification helps the model to focus on objects.


\subsection{Cortical Magnification induces a trade-off between central and peripheral visual content}

Our previous analysis suggests that making objects appear larger is key to the performance of Cortical Magnification. However, a smaller crop of the image can also achieve a similar effect. In this section, we analyze the impact of the central crop size (cf. \sectionautorefname~\ref{sec:datasets}) on object learning with and without Cortical Magnification. Specifically, in the Toddler Fixation dataset, we used crop sizes of $64\times64$, $128\times128$ (the same size used in other experiments), and $240\times240$, and then uniformly resized them to $128\times128$. In the CORe50 dataset, the original images have a size of $350\times350$ pixels, from which we crop at various sizes from the center of the original image.

In \figureautorefname~\ref{fig:acc_hycor}A, we observe that the performance of Cortical Magnification highly depends on the crop size. However, applying Cortical Magnification (dashed lines) yields higher accuracy than cropping alone (solid lines) for crop sizes higher than $64\times64$. The highest recognition accuracy is achieved at $128\times128$, presumably because this crop size best fits the object. We interpret this result as showing that Cortical Magnification leads to a better trade-off between extracting features in the periphery versus in central vision, which is impossible for a simple cropping method.

\subsection{The Effect of Cortical Magnification Parameters}
\begin{figure}[ht]
  \centering
  \includegraphics[width=1\linewidth]{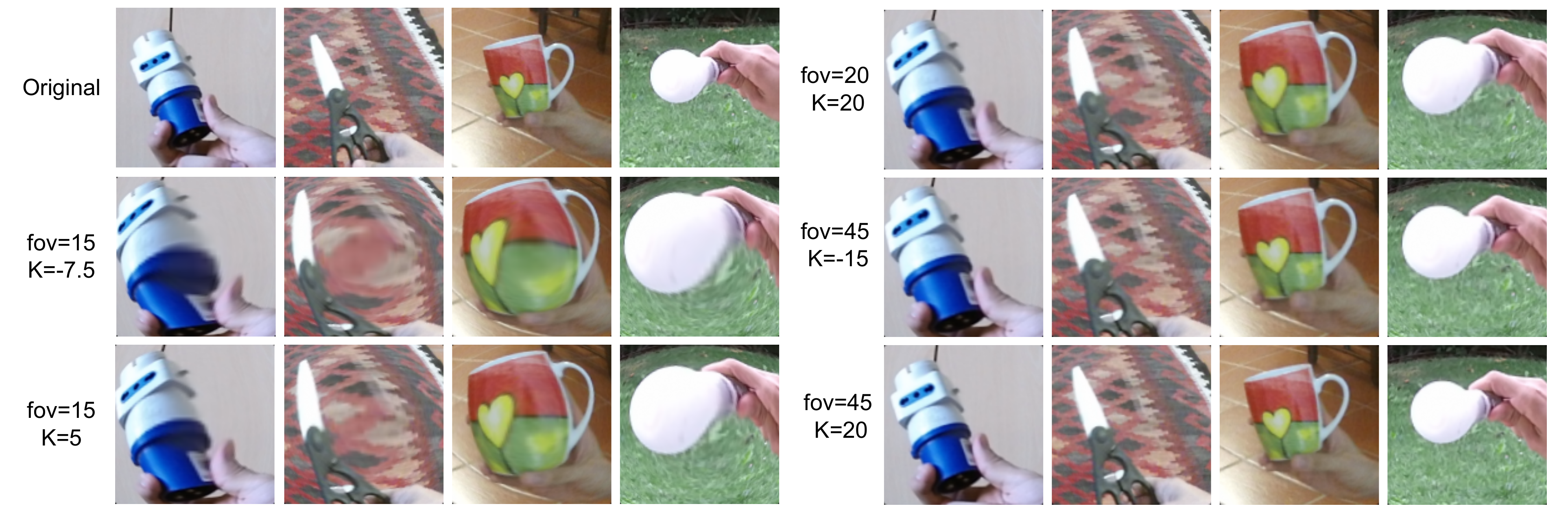}
  \caption{The visual effect of different Cortical Magnification hyperparameters. In the experiment, the parameters are set as $fov \in \{15, 20, 45\}$ and $K \in \{-15, -7.5, 5, 20, 40, 50\}$.}
  \label{fig:hycor}
\end{figure}

\begin{figure}[t]
  \centering
  \includegraphics[width=1\linewidth]{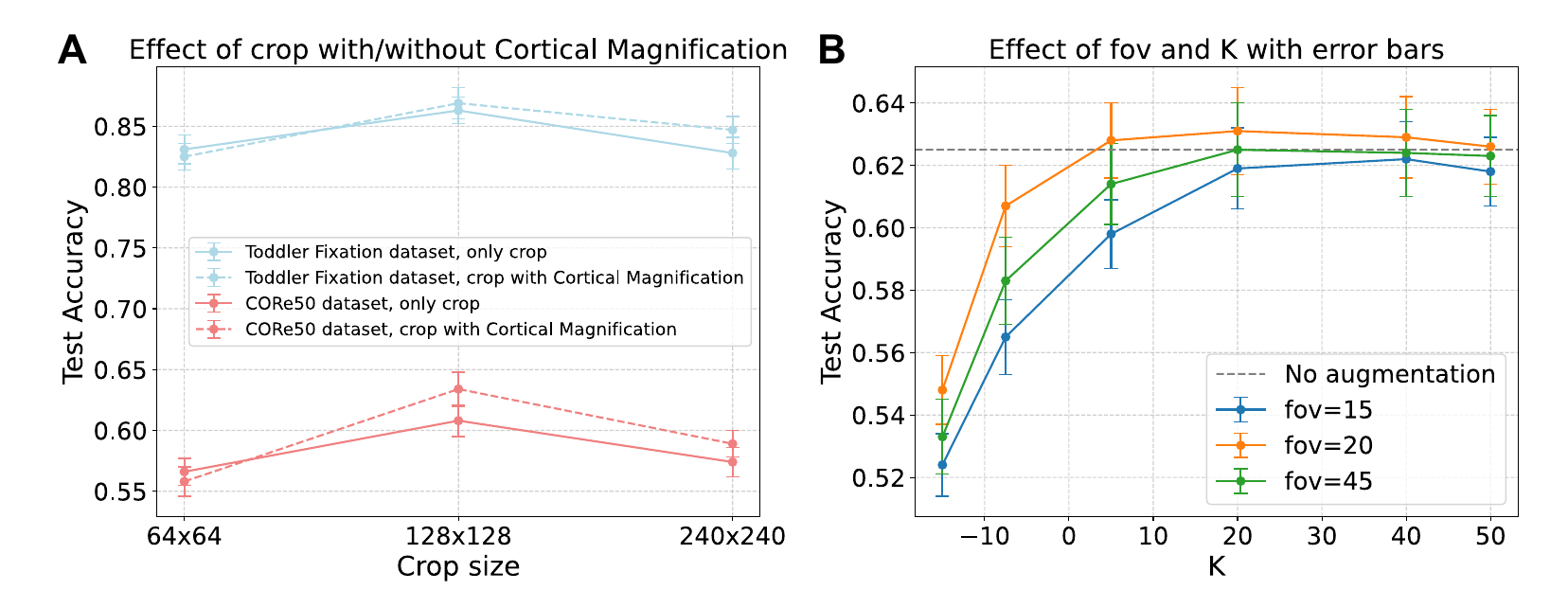}
  \caption{{\bf (A).} Effect of Crop Size with or without Cortical Magnification. Blue and red represent the Toddler Fixation dataset and the CORe50 dataset, respectively. Solid and dashed lines indicate using only cropping and cropping followed by Cortical Magnification. {\bf (B).} Effect of $fov$ and $K$. A horizontal dashed line indicates the baseline test accuracy achieved without any augmentation. Error bars represent the standard deviation across three runs with different random seeds.}
  \label{fig:acc_hycor}
\end{figure}

In this section, we analyze the effects of $fov$ and $K$ hyperparameters, which rule the level of distortion of the image, on the learning process. \figureautorefname~\ref{fig:hycor} demonstrates how varying the hyperparameters affects the visual appearance of the input images. We evaluate SimCLR-TT with a ResNet18 encoder trained on half of the CORe50 dataset. In \figureautorefname~\ref{fig:acc_hycor}B, we find that the best recognition accuracy occurs at $K=20$, $fov=20$. A larger $fov$, like $45$, maintains accuracy close to the original image, while smaller values (e.g., $fov=15$) introduce excessive distortion, impairing learning. Similarly, larger $K$ values preserve object features, while very small $K$ lead to severe distortions and accuracy drops. Therefore, effective representation learning benefits from moderate distortion.




\section{Discussion}
We investigated the impact of the varying resolution of human vision on the learning of object representations. To model this varying resolution, we have incorporated two image augmentations, ``Foveation'' that blurs the image towards the periphery, and ``Cortical Magnification'' that simulates a fish-eye view. Then, we trained bio-inspired visual learning \textcolor{black}{models} on top of visual sequences capturing the visual experience of humans during interactions with objects. Our findings show that the Cortical Magnification augmentation \textcolor{black}{consistently improves} the learning of object representations. However, the Foveation augmentation leads to mixed results. Our analysis suggests that the performance of Cortical Magnification derives from making objects appear larger. We also found that Cortical Magnification induces a better trade-off, compared to keeping only central vision, between important central vision and slightly less important periphery. Overall, our results suggest that the foveated nature of human vision may play an important role during the learning of object representations.


While our model captures the feature of cortical magnification, it does not fully replicate the complex retinotopic warping or individual variability observed in biological systems \cite{dumoulin2008population}. Instead, we aimed to \textcolor{black}{use} a computationally efficient approximation that emphasizes central object features and down-weights distracting background information. As our results show, this relatively coarse approximation already induces beneficial effects for self-supervised object learning by making objects appear larger and enhancing object-centric feature extraction, consistent with prior findings that \textcolor{black}{objects filling more of the learner’s visual field} facilitate representation learning \cite{yu2024active,orhan2024learning}.

However, the backbone networks are not intended to be fully detailed or biologically accurate models. They serve as practical approximations that enable effective self-supervised representation learning. Future work could explore more biologically detailed architectures incorporating recurrent or predictive \textcolor{black}{processing} mechanisms, to further narrow the gap \textcolor{black}{to} biological visual processing and learning.




\bibliographystyle{IEEEtran}
\bibliography{bib}

\end{document}